# Corrosion Detection for Industrial Objects: From Multi-Sensor System to 5D Feature Space

D. Haitz*, B. Jutzi, P. Hübner, M. Ulrich

Institute of Photogrammetry and Remote Sensing, Karlsruhe Institute of Technology (KIT), Karlsruhe, Germany
(dennis.haitz, boris.jutzi, patrick.huebner, markus.ulrich)@kit.edu

**Commission I, WG I/6**

**ABSTRACT:**

Corrosion is a form of damage that often appears on the surface of metal-made objects used in industrial applications. Those damages can be critical depending on the purpose of the used object. Optical-based testing systems provide a form of non-contact data acquisition, where the acquired data can then be used to analyse the surface of an object. In the field of industrial image processing, this is called surface inspection. We provide a testing setup consisting of a rotary table which rotates the object by 360 degrees, as well as industrial RGB cameras and laser triangulation sensors for the acquisition of 2D and 3D data as our multi-sensor system. These sensors acquire data while the object to be tested takes a full rotation. Further on, data augmentation is applied to prepare new data or enhance already acquired data. In order to evaluate the impact of a laser triangulation sensor for corrosion detection, one challenge is to at first fuse the data of both domains. After the data fusion process, 5 different channels can be utilized to create a 5D feature space. Besides the red, green and blue channels of the image (1-3), additional range data from the laser triangulation sensor is incorporated (4). As a fifth channel, said sensor provides additional intensity data (5). With a multi-channel image classification, a 5D feature space will lead to slightly superior results opposed to a 3D feature space, composed of only the RGB channels of the image.

## 1. INTRODUCTION

Within the process of industrial production, image processing is often connected to quality control of manufactured parts. While there are other industrial applications for image processing, e. g. completeness checking, position detection or object identification (Steger et al., 2018), our work focuses on the task of surface inspection of metal-made industrial objects. While there is a number of types of anomalies that can occur on the surface of such objects, e. g. deformation or scratches, we are interested in the detection of corrosion exclusively. Further on, the objects to be tested are of rotation symmetric shape, specifically of cylindric shape.

Our testing system consists of a rotary table (Figure 1), which rotates the object by 360 degrees. Within this process, a set of optical sensors acquire data of the object. The sensor system consists of active optical sensors, which are based on the laser triangulation measurement principle in order to acquire range data. A laser-beam projects a light plane onto the object, so that the intersection of this plane results in a scattered light on the object. A camera, which is integrated within the same hardware system as the laser projector, then captures the scattered light. In order to calculate the distance to the laser point by triangulation, the system must be calibrated, i. e. the interior orientation of the camera and the relative orientation of the camera and laser projector must be known. This type of sensor is also called sheet of light sensor (Steger et al., 2018). Apart from range measurement, the sensor is also capable of acquiring intensity data, which is a measurement of the light of the laser-beam, reflected by the object (LMI Technologies, 2021).

The other type of sensor used in our system are industrial RGB cameras for the acquisition of three channel images. These cameras will be referred to as passive sensors within this contribution. The data acquisition process is split up for each sensor, so that the acquisition takes place sequentially for each of the two sensors.

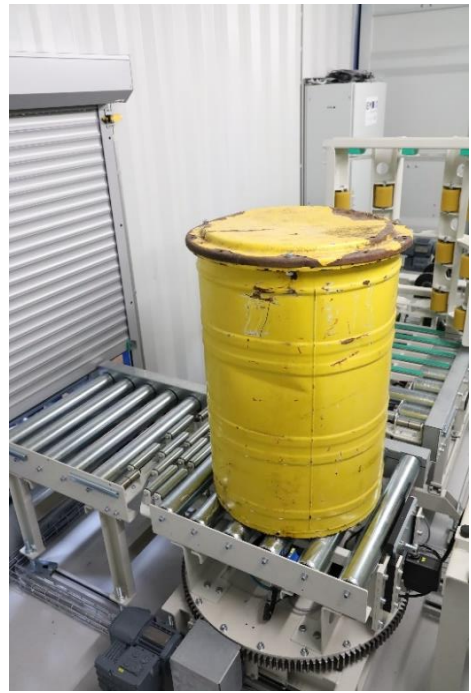

**Figure 1.** Our testing object, a yellow barrel, located on the rotary table.

For the task of corrosion detection, the data acquired by the two sensors needs to be fused, so that a multi-channel image can be generated for the task of corrosion detection, which is realized by a pixel-wise supervised classification.

---

* Corresponding author

The first part of our contribution is a measurement setup consisting of a rotary table and a multi-sensor system, which acquires data from rotation symmetric, cylindric-shaped objects, located on the rotary table. After the acquisition, the second part of our contribution refers to the fusion of the data acquired by the active and passive sensors. This leads to a five-channel image, which consists of the three channels $I_{RGB}$, acquired by the passive sensor, as well as range data $R$ and intensity data $R_I$, acquired by the active sensors. The last part of our contribution contains a supervised classification in order to detect corrosion on the surface of the object. In order to determine the influence of each sensor, classification and the evaluation of the classification result is performed and compared for different channel-combinations. Our contribution consists of:

- Data acquisition of rotation symmetric objects located on a rotary table, using active and passive sensors
- Data augmentation as an additional image source
- Data fusion to obtain multi-channel images
- Corrosion detection through supervised classification
- Evaluation of the impact of the active sensors for corrosion detection

Our paper is structured as follows. In Chapter 2, related work is depicted. Chapter 3 contains a detailed description of the data fusion process as well as the classification step. This is followed by Chapter 4, where the data acquisition will be explained. Results of the classification are shown in Chapter 5 by showing classifier images and common classification metrics. Thereafter, the obtained and shown results are discussed. An outlook on future work is presented in the last chapter.

## 2. RELATED WORK

The topic of our paper is related to a number of different research fields such as the registration of multi-sensor data of different modalities, corrosion detection and the inspection of cylindrical objects in industrial settings. This section presents a short overview of related work on the mentioned topics.

The task of fusing multi-modal data from different sensors is a broad research field in itself (Sasiadek, 2002). In the case of active or passive imaging sensors, this task can either be approached by determining the relative poses between the different sensors or by directly registering the resulting data without explicit regard for sensor setup (Weinmann et al., 2011, 2012). In static sensor setups, the relative poses between the involved sensors can be determined in an offline phase by means of calibration procedures (Avetisyan et al., 2014; Steger et al., 2018; Usamentiaga and Garcia, 2019). Dynamic sensor arrangements however, require continuous real-time tracking of the respective relative poses (Hübner et al., 2018).

Concerning our specific sensor setup, the task at hand is to fuse data of an ordinary perspective color camera with an active sensor providing intensity and range values for one-dimensional profile lines. Again, this task of combining perspective images with line scans can be approached be calibrating the relative pose between the sensors (Hu et al., 2016; Hillemann and Jutzi, 2017) in a dedicated calibration phase with specific calibration targets. Other approaches focus on refining an initially calibrated relative pose by taking into account the recorded scene data of both sensors (Jutzi et al., 2014; Bybee and Budge, 2019). Due to our specific measurement setup, is feasible for us to directly register stitched images of both sensors without relying on their relative pose.

Regarding our application context, the automated visual detection of corrosion is an active field of research as well (Ahuja and Shukla, 2017). Some approaches focus on detecting corrosion from images via the geometric properties of rust spots on otherwise more or less homogeneous surfaces (Wang and Cheng, 2016; Enikeev et al., 2017). Another approach focuses on radiometric properties and aims to detect corrosion pixels via their color while investigating the impact of different color space transformation in this context (Chen et al., 2009). Yet others use texture properties (Medeiros et al., 2010; Feliciano et al., 2015) or follow a combined approach using color properties as well as shape and texture for corrosion detection (Choi and Kim, 2005; Bondada et al., 2018). Recently, corrosion detection approaches based on deep learning methods are gaining in prevalence, for instance as patch-based binary classification regarding the presence of corrosion applied in a sliding-window manner (Atha and Jahanshahi, 2018; Papamarkou et al., 2021) or as pixel-wise semantic segmentation (Liu et al., 2018; Duy et al., 2020; Fondevik et al., 2020; Rahman et al., 2021).

Finally, related work concerning the measurement setup for the automated inspection of cylindrical or rotationally symmetric objects is addressed. Proposed sensor setups to this aim for instance rely on ordinary cameras (Ayub et al., 2014; František et al., 2018; Govindaraj et al., 2018; Cerezci et al., 2020; Schlagenhauf et al., 2021), line cameras (Zographos et al., 1997) or high-speed cameras (Tabata et al., 2010) for laterally observing a rotating object. While the inspection of the object surface is often conducted in a per-frame manner, some approaches aim at aggregating the individual observations during the rotation of the object to a complete panorama of the lateral object surface like we do (Cerezci et al., 2020; Schlagenhauf et al., 2021). However, to the best of our knowledge, we are the first to use multiple sensors with different modalities in this setting. Besides rotating a cylindrical object while laterally observing its surface, some other approaches for instance use a setup with a single camera and mirrors in order to observe the whole lateral object surface at once (Ali and Lun, 2019) or observe the object in top-down view (Chen et al., 2020) when only its contour is of interest.

## 3. METHODOLOGY

In this chapter, the data fusion as well as the classification are described. The data acquisition is described separately in Chapter 4. So, the precondition of this chapter is already acquired data, which needs to be fused to obtain n-channel images. For our experiments, we only acquire data from the coat of the object.

### 3.1 Data Fusion

For the data fusion step, the data needs to be preprocessed beforehand. These steps are described separately for each the active and passive sensor in the following sections.

**Active Sensor**
The data of the active sensor consists of so-called line profiles, where each profile itself consists of a few thousand, equidistant measurement points. Each point consists of an x- and z-coordinate, as well as an intensity value. The z-coordinate contains range information, whereas the x-coordinate denotes the position of the point along the direction of the projected line. In addition, the intensity value contains the reflectivity of the laser-beam as an unsigned 8-bit value. While the object rotates by 360 degrees, the sensor acquires line profiles with a measurement frequency of 250 Hz. With a *de facto* constant rotation speed, a full rotation of 360 degrees takes 42 seconds. The resulting number of line profiles is 10542, where each line profile contains

3600 measurement points. In a first step, a range map image from the z-coordinates in the line profiles, as well as an intensity map from the intensity values, are created. This is done by concatenating all line profiles, which results in images with a width of 10542 px and a height of 3600 px. A further manual step is to crop the object in both images to remove background, which reduces the height to 2336 px. In order to normalize the range data, every column in the image, which originally stems from the line profile, is subtracted by its median. After the subtraction, the column gets scaled to the interval [0, 65535], which is the value range of the unsigned 16-bit integer data type for the subsequent computation purposes.

**Passive Sensor**

The passive sensor acquires 1461 images within one full rotation of 360 degrees. From each image, a crop of the height of the object and width of 7 px is taken and concatenated with the previous crop. The width of 7 px is determined in an empirical way. After concatenation of the single image crops, the image of the coat of the object has a width of 8372 px and a height of 3062 px.

**Image Stacking**

In order to stack the images from the previous steps to a multichannel image, a resampling must be applied, so that every image has the same geometrical resolution. The images from the active sensor – $R$ and $R_I$ – are resampled to align with $I_{RGB}$ of the passive sensor. As can be seen in Figure 3a, $R$ and $R_I$ of the active sensor do not align with $I_{RGB}$ of the passive sensor very well, so a further image transformation is needed. A possible solution is to estimate a homography by using an Interest-Operator, e. g. SIFT (Lowe, 2004) or Förstner-Points (Förstner and Gülch, 1987), and find a robust mapping from $R_I$ to $I_{RGB}$. However, the images still do not align well enough at the upper and lower border regions of the object. These regions are of high interest, because most of the corrosion of the object can be found there. Our solution is to utilize Förstner-Points as Interest-Points in both $R_I$ and $I_{RGB}$ to find correspondences in both images within a certain positional tolerance. For each two corresponding points, the difference in x- and y-direction is calculated. By utilizing a thin-plate spline interpolation (Bookstein, 1989), a vector field for both the x- and y-direction in the form of a grayscale image is generated. Both vector fields serve as the input for an image warping, which outputs the locally rectified $R_I$ w. r. t. $I_{RGB}$. The resulting transformation rule is applied to $R$, too, in order to obtain the same geometrical resolution.

**3.2 Data Augmentation**

In order to be able to utilize the range data as an indicator for the detection of corrosion, the differences of range are rather of interest than the range values themselves. To obtain those differences, we calculate the first derivative of $R$ by applying a sobel image filter (Gonzales and Woods, 2018). As a result, each pixel in the sobel image represents the local difference in height. This resulting image $R_S$ (Figure 2b) acts as our actual range image in the multi-channel image.

Another augmentation strategy is to change the color space from RGB (Figure 2a) to HSI (Figure 2d-2f). As described in chapter 2, some research on the topic of image-based corrosion detection has shown, that the HSI color space might have some advantages over the RGB image space when it comes to applying a classification algorithm (Choi and Kim, 2005).

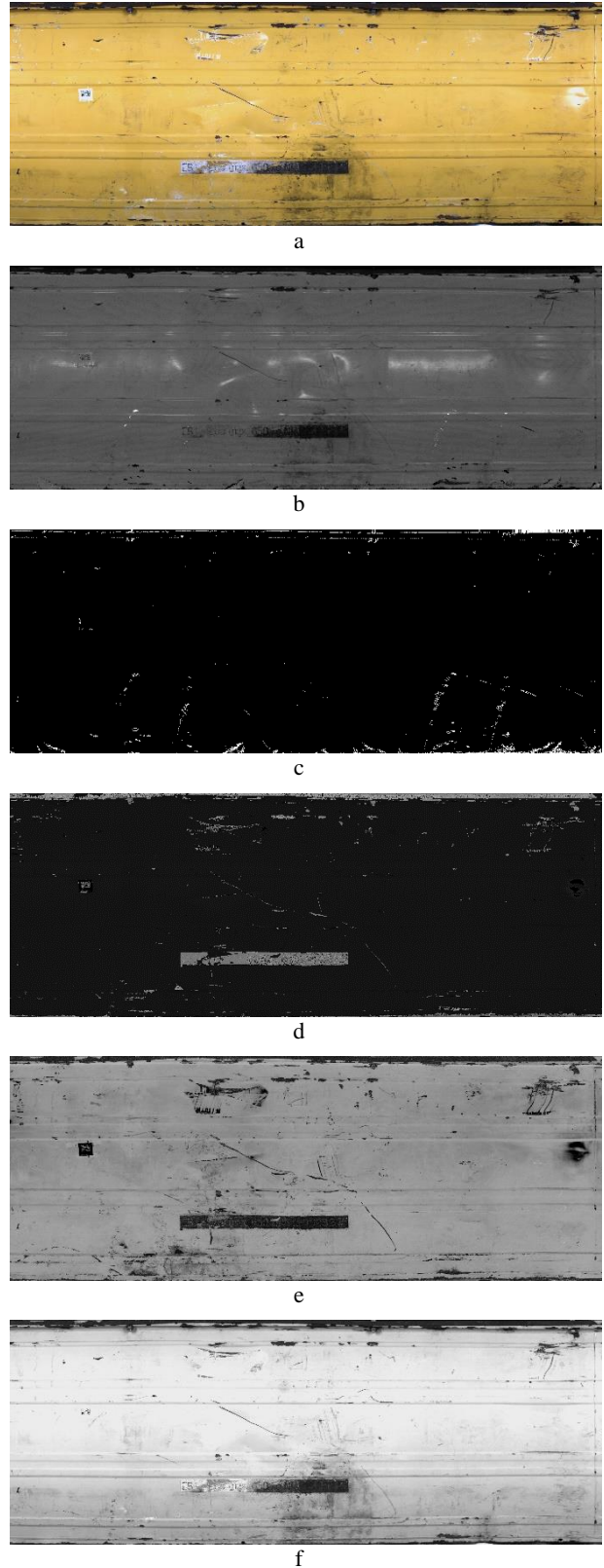

**Figure 2.** The eight utilized image channels: a) Three channels RGB, b) Range Intensity, c) Range derivative from sobel operation, d) HSI-Hue, e) HSI-Saturation, f) HSI-Intensity.

## 3.3 Classification

After applying the previous augmentation methods, the next step is to detect corrosion by applying a per-pixel classification. Because we are doing supervised classification, at first training and test datasets need to be created. This is done by marking polygons on the image stack, which belong to a certain object class. The classification scheme is derived from the objects that are visible on the surface. This results in the following seven classes to be separated:

- Unaffected surface
- Corrosion
- Red lacquer
- White lacquer
- Light dirt
- Dark dirt
- Reflection

Because our goal is to only detect corrosion, the classes are then aggregated into a binary classification result, where the first class is corrosion and the second class is non-corrosion or unaffected, respectively. The classification with seven classes is done to overclassify the image with the intention to reduce overlapping of feature vectors belonging to different classes within the feature space. For our experiments, we use a Random Forest classifier (Breiman, 2001) only, because we are only interested in the evaluation of sensor types at this stage of our work. We chose the Random Forest classifier because it is a proven classification algorithm for the task of per-pixel classification. For our experiments, we use one set of hyperparameters. Specifically, those are 128 decision trees, each with a depth of 20.

For training and evaluating the classifier, we use between-class-balanced datasets. Usually, balanced datasets are preferred to imbalanced datasets for classification if possible. This is due to the fact that classification algorithms yield better results with balanced datasets (Ng et al., 2015). Our datasets in their original form are imbalanced, because some class instances only appear in small amounts on the surface of the object, whereas other instances appear in higher numbers. In order to balance the dataset, we apply the undersampling technique, which randomly removes data from all class-samples apart from the one with the least amount of data until every class has the same amount of data (He and Garcia, 2009).

The classification results are evaluated by utilizing the widely used classification metrics precision, recall, f1-score and overall-accuracy (Table 1 and 2), which are derived from a confusion matrix (Prowers, 2007). Precision shows the false positive rate, which is an indicator of how much data of other classes was classified to a specific class. Recall is an indicator of how much data was falsely classified and put to the wrong class. The f1-score is the harmonic mean of precision and recall. The overall accuracy shows the performance over all classes by summing up the true positives and divide them by the number of all data.

Further on, we apply the classification to different channel-combinations. The following channel-combinations are used for classification and evaluation:

- $I_{RGB}$
- $I_{RGB}, R_I$
- $I_{RGB}, R_I, R_S$
- $I_{RGB}, R_I, R_S, I_{HSI}$
- $I_{HSI}$
- $I_{HSI}, R_I$
- $I_{HSI}, R_I, R_S$

| Channels | Classes | Metrics | | | OA |
|---|---|---|---|---|---|
| | | P | R | F1 | |
| $I_{RGB}$ | Unaffected | 0.93 | 0.85 | 0.89 | 0.79 |
| | Corrosion | 0.94 | 0.78 | 0.85 | |
| | R. Lacquer | 0.87 | 0.66 | 0.75 | |
| | W. Lacquer | 0.63 | 0.92 | 0.75 | |
| | Light Dirt | 0.70 | 0.56 | 0.62 | |
| | Dark Dirt | 0.70 | 0.91 | 0.79 | |
| | Reflection | 0.91 | 0.86 | 0.88 | |
| $I_{RGB}, R_I$ | Unaffected | 0.93 | 0.89 | 0.91 | 0.82 |
| | Corrosion | 0.93 | 0.81 | 0.87 | |
| | R. Lacquer | 0.83 | 0.90 | 0.86 | |
| | W. Lacquer | 0.66 | 0.89 | 0.75 | |
| | Light Dirt | 0.70 | 0.58 | 0.63 | |
| | Dark Dirt | 0.85 | 0.83 | 0.84 | |
| | Reflection | 0.92 | 0.87 | 0.90 | |
| $I_{RGB}, R_I, R_S$ | Unaffected | 0.93 | 0.86 | 0.90 | 0.81 |
| | Corrosion | 0.93 | 0.83 | 0.88 | |
| | R. Lacquer | 0.81 | 0.88 | 0.84 | |
| | W. Lacquer | 0.64 | 0.90 | 0.75 | |
| | Light Dirt | 0.71 | 0.55 | 0.62 | |
| | Dark Dirt | 0.83 | 0.81 | 0.82 | |
| | Reflection | 0.91 | 0.87 | 0.89 | |
| $I_{RGB}, R_I, R_S, I_{HSI}$ | Unaffected | 0.94 | 0.86 | 0.90 | 0.81 |
| | Corrosion | 0.94 | 0.79 | 0.85 | |
| | R. Lacquer | 0.82 | 0.86 | 0.84 | |
| | W. Lacquer | 0.64 | 0.89 | 0.74 | |
| | Light Dirt | 0.68 | 0.56 | 0.62 | |
| | Dark Dirt | 0.81 | 0.82 | 0.82 | |
| | Reflection | 0.92 | 0.87 | 0.89 | |
| $I_{HSI}$ | Unaffected | 0.93 | 0.87 | 0.89 | 0.79 |
| | Corrosion | 0.92 | 0.78 | 0.84 | |
| | R. Lacquer | 0.87 | 0.65 | 0.75 | |
| | W. Lacquer | 0.64 | 0.94 | 0.76 | |
| | Light Dirt | 0.71 | 0.55 | 0.62 | |
| | Dark Dirt | 0.70 | 0.91 | 0.79 | |
| | Reflection | 0.91 | 0.85 | 0.88 | |
| $I_{HSI}, R_I$ | Unaffected | 0.93 | 0.87 | 0.90 | 0.81 |
| | Corrosion | 0.94 | 0.81 | 0.87 | |
| | R. Lacquer | 0.82 | 0.89 | 0.86 | |
| | W. Lacquer | 0.64 | 0.87 | 0.73 | |
| | Light Dirt | 0.68 | 0.56 | 0.61 | |
| | Dark Dirt | 0.83 | 0.82 | 0.83 | |
| | Reflection | 0.92 | 0.87 | 0.89 | |
| $I_{HSI}, R_I, R_S$ | **Unaffected** | **0.93** | **0.86** | **0.89** | **0.82** |
| | **Corrosion** | **0.94** | **0.84** | **0.89** | |
| | **R. Lacquer** | **0.82** | **0.87** | **0.84** | |
| | **W. Lacquer** | **0.64** | **0.90** | **0.75** | |
| | **Light Dirt** | **0.72** | **0.56** | **0.63** | |
| | **Dark Dirt** | **0.82** | **0.82** | **0.82** | |
| | **Reflection** | **0.93** | **0.87** | **0.89** | |

**Table 1.** Evaluation metrics for the classification of seven classes for each channel-combination. The metrics are precision (P), recall (R), f1-Score (F1) and overall accuracy (OA).

## 4. DATA ACQUISITION

In this chapter, the setup for the data acquisition is described. There are two main components in our setup: The rotary table where the object is located and the multi-sensor system, consisting of active and passive sensors.

The rotary table is part of a conveyor belt. The task of the conveyor belt is to move the object to every testing step, where data of each element of the object is acquired. In our specific case, the object is a cylindric barrel, from which data is acquired individually for the top, coat and bottom of the barrel. At this stage of our work, we only focus on the coat of the barrel. When the object arrives on the rotary table, it will be rotated by 370 degrees. This is because the first 5 degrees of the rotation are a buffer for the rotary motor to accelerate, and the last 5 degrees are a buffer for the motor to slow down again, so that during a full rotation of 360 degrees the rotation speed remains constant. For the data acquisition, two rotation steps are taken for each the active and passive sensor. The reason therefor is that for the active sensor, the measurement quality is higher with low illumination, whereas the passive sensors need fairly high illumination. Both sensors are mounted on the side of the inner wall of the testing system. The laser beam of the active sensor is projected along the vertical axis of the object. After the rotation table reaches constant speed, the active sensor receives a trigger signal in order for the sensor to start data acquisition. The acquisition frequency is 250 line profiles per second. Every line profile consists of 3600 measurement points, where every point contains an x-coordinate, which denotes the position along the laser-beam, a z-coordinate, which is the range data and an intensity value, which is a measurement of reflection of the laser-beam at that very measurement point.

The data from the passive sensor are RGB images, which are acquired in a second, full rotation after the acquisition of the active sensor has ended. After the triggering of the passive sensor, the acquisition rate is the maximum possible frame rate of the sensor, which is 31 frames per second.

Our object is of rotation symmetric, cylindric shape. A major challenge that arises within the process of data acquisition is the offset of the axis of the cylinder to the rotation axis of the rotation table. This offset leads to an additional movement in z-direction (range) as well as in y-direction (time). In order to equalize the range data, the median subtraction described in Chapter 3.1 is applied.

## 5. RESULTS

In this chapter, the results of the data fusion, data augmentation and classification are presented. The data fusion is presented showing $I_{RGB}$ converted to a grayscale image, stacked with $R_I$ for a two-channel false-color representation (Figure 3). This way, the surface of the object gets visualized at first in its unaligned stage, then after the application of the fusion process in its aligned stage. The augmentation process is already shown in Figure 2 and consists of the derivative of the range image as well as the generation of $I_{HSI}$ by applying a color space transformation to $I_{RGB}$. The classification results are represented as a seven-class segmentation of the multichannel images (Figure 4a). Other than that, the classification metrics precision, recall, f1-score and overall accuracy are shown for each channel-combination (Table 1). For the final two-class representation, all classes apart from the corrosion class are merged to an unaffected class. The result is also shown as a segmented image (Figure 4b) as well as the aforementioned evaluation metrics (Table 2).

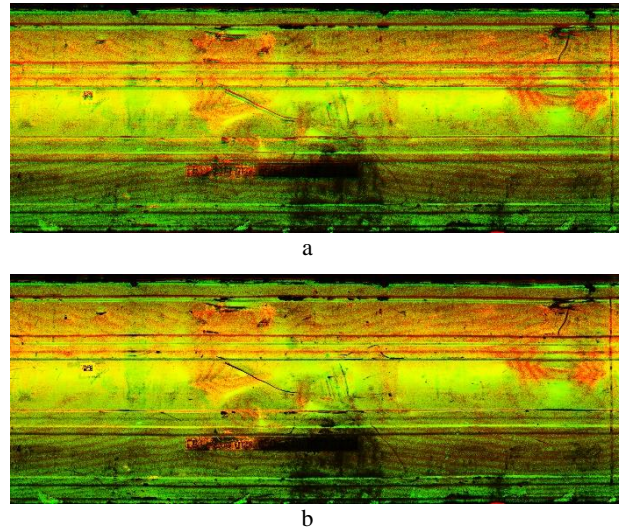

**Figure 3.** Three channel images with the RGB image converted to grayscale as the red channel, Range Intensity as the green channel and a zero image as the blue channel: a) Shows the state of the not transformed intensity image while b) shows the application of the thin-plate-spline-interpolation based transformation.

## 6. DISCUSSION

In this chapter, the previously shown results are discussed.

### 6.1 Data acquisition

The data acquisition process is straight forward in the sense, that data is captured by active and passive sensors while the object is rotated by 360 degrees. The rotatory table holds a speed, which can be considered constant and mechanical issues which could interfere are not known at this point. Because it is not possible to align the axis of the cylindric object with the rotation axis of the rotary table, the resulting eccentricity is still a challenge for which a solution is needed. The movement resulting from the aforementioned eccentricity in depth as well as the time axis leads to different measurement resolutions. This problem deconvolves in the data fusion process on the one hand, where data from the 2D and 3D domain is stacked to a multi-channel image. On the other hand, a qualitative statement of the amount of corrosion on the surface is getting harder to determine, if the measurement resolution differs throughout the acquisition process.

| Channels | Classes | Metrics | | | |
|---|---|---|---|---|---|
| | | P | R | F1 | OA |
| $I_{RGB}$ | Unaffected | 0.96 | 0.99 | 0.98 | 0.96 |
| | Corrosion | 0.94 | 0.78 | 0.85 | |
| $I_{RGB}$, $R_I$ | Unaffected | 0.97 | 0.99 | 0.98 | 0.96 |
| | Corrosion | 0.93 | 0.81 | 0.87 | |
| $I_{RGB}$, $R_I$, $R_S$ | Unaffected | 0.97 | 0.99 | 0.98 | 0.97 |
| | Corrosion | 0.93 | 0.83 | 0.88 | |
| $I_{RGB}$, $R_I$, $R_S$, $I_{HSI}$ | Unaffected | 0.97 | 0.99 | 0.98 | 0.96 |
| | Corrosion | 0.94 | 0.79 | 0.85 | |
| $I_{HSI}$ | Unaffected | 0.96 | 0.99 | 0.98 | 0.96 |
| | Corrosion | 0.92 | 0.78 | 0.84 | |
| $I_{HSI}$, $R_I$ | Unaffected | 0.97 | 0.99 | 0.98 | 0.97 |
| | Corrosion | 0.94 | 0.81 | 0.87 | |
| $I_{HSI}$, $R_I$, $R_S$ | **Unaffected** | **0.97** | **0.99** | **0.98** | **0.97** |
| | **Corrosion** | **0.94** | **0.84** | **0.89** | |

**Table 2.** Aggregation of all non-corrosion classes to one class and evaluation of the resulting classes unaffected and corrosion. The metrics are precision (P), recall (R), f1-Score (F1) and overall accuracy (OA).

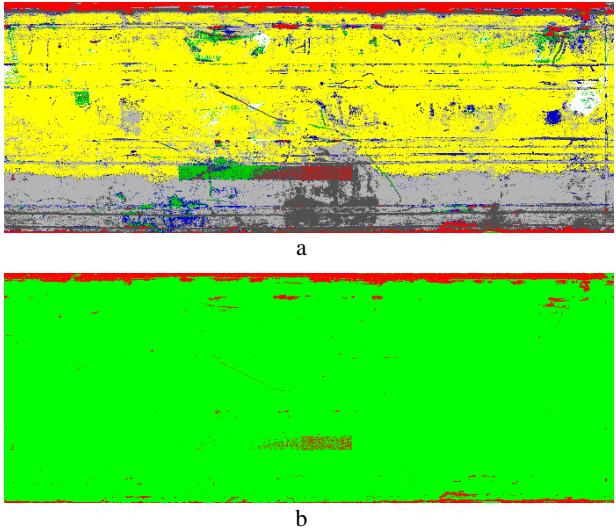

**Figure 4.** Results for the highest classification performance w. r. t. corrosion detection for seven and two classes. a) Shows the seven class result for the HSI, Range Intensity and Range Derivative image with the following pixel colors per class: yellow – unaffected, red – corrosion, blue – red lacquer, green – white lacquer, light gray – light dirt, dark gray – dark dirt, white – reflection. b) shows the two class result for the HIS, Range Intensity and Range Derivative image with green pixels representing the unaffected class, and red pixels the corrosion class.

### 6.2 Data augmentation

The color space transformation from $I_{RGB}$ to $I_{HSI}$ leads to a representation, which yields the advantage to classify data independent of the surface color by leaving out the hue channel. Channel-combinations integrating $I_{HSI}$ also tend to lead to similar, if not slightly higher evaluation metrics in comparison to $I_{RGB}$. The sobel filtered Range image $R_S$ has a very low value range as can be seen in Figure 2c. Further experiments like scaling the value range or investigating a channel-combination of $I_{RGB}$ and $R_S$ might lead to insights, whether this channel is useful for the detection of corrosion.

### 6.3 Data fusion

The data fusion at this point is done manually to some extent. There are three steps that are taken for the data fusion:

- Manual preprocessing of the data of both sensors
- Geometrical transforming the data of the active sensor to obtain the geometrical resolution of the data of the passive sensor
- Stacking the data of both sensors to multi-channel images

The pre-processing in the domain of the passive sensors consists of cutting out rectangles in every acquired image at the objects center, each of the height of the object and a width of 7 px. The resulting rectangles are then concatenated. Because of the movement of the object described in Chapter 6.1, a static width does not lead to an optimal visual representation of the coat of the object. Especially the boarders of diagonal structures exhibit spike-like patterns, which actually should be homogenous. The cylindrical geometry of the object as well as the perspective distortion is not taken into account at this stage. Intensity and range images $R_I$ and $R$ are created by concatenating the acquired line profiles along the time-axis. The resulting images are then cropped manually in order to only contain the object. Then following, geometrical transformations are applied. At first, $R_I$ and $R$ are resized in the lateral direction. The transformation based on the thin-plate splines-interpolation warps the image based on the density and spatial tolerance of corresponding interest points in both $I_{RGB}$ and $R_I$. This is the crucial step of the geometrical transformation, because it determines how well the images of both domains are aligned. The result of this transformation step is a quite usable basis in our context of determining the impact of feature space augmentation. However, some regions in both domains do not align very well. Especially the upper border of the object still shows some offsets regarding the structures that appear on the surface of the object. This might be caused due to a low number of corresponding interest points in both images in that region. In order to classify relative height differences, a sobel filter is applied to $R$. The resulting $R_S$ is integrated in the multi-channel images used in the following classification.

### 6.4 Classification

We perform per-pixel-classifications for every channel-combination using a Random Forest classifier with the same hyperparameters for every training. The training and test areas are left the same for the training and evaluation of each stack. The evaluation metrics in Table 1 for the seven classes show that there is no significant difference between the different channel-combinations regarding the performance of the classifier. Corrosion as the most important class shows high values between 0.92 and 0.94 in precision across all stacks. This shows that the data of the other classes do not have a significant negative impact on the classification result. Recall values differ to some extent across the channel-combinations with the lowest number of 0.78 for $I_{RGB}$ and $I_{HSI}$ to 0.84 for $I_{HSI}$, $R_I$, $R_S$. A higher recall means that actual data for the corrosion is not falsely classified to another class. While 0.84 is an acceptable value for classification results, it is still desirable to optimize the classification process. With a precision value of 0.94, among other stacks the before mentioned $I_{HSI}$, $R_I$, $R_S$ stack performs best in our experiments. Prominent examples of false positives can be seen in Figure 4b, where in the center on the lower part of the image an almost rectangular structure is falsely detected as corrosion (red color). This means that even though the precision as an indicator of the impact of false positives is quite high, an optimization of the classification still needs to be considered.

### 6.5 Impact of the active sensor

As discussed in the previous section, a 5D feature space yields the best results. However, the differences in evaluation metrics between the channel-combinations are not highly significant. By comparing the class metrics across all channel-combinations, only in some minor cases significant differences can be observed, e. g. the quite low recall of the Red Lacquer class in $I_{RGB}$ and $I_{HSI}$ on the one hand, and the much higher recall of said class for the $I_{RGB}$, $R_I$ stack.

## 7. CONCLUSION

In our contribution, we presented a testing setup for cylindrical objects consisting of a rotary table and a multi-sensor system in order to detect corrosion on the surface of an industrial object.

The eccentricity obtained by the offset of the cylindrical axis of the object and the rotation axis of the rotation table leads to challenges in the data processing for the sake of unwinding the objects coat in a geometrically exact manner. Another challenge is the data fusion, which at this point is done with highly structured images, e. g. surface scratches or corrosion spots. The data fusion is based on finding correspondences in both the active and passive sensor domain. However, objects with no such structure have to be considered for future work. So other methods need to be evaluated in order to perform data fusion.

Our main goal was to show if the augmentation of the feature space provides an advantage over the usage of data of only one sensor. A 5D feature space consisting of a HSI image, intensity image obtained from the active sensor as well as the range differences of the active sensors showed best results. However, these results are only marginally superior to RGB or HSI images. This leads to the question, if the data of the active sensor should be considered under the condition of a fairly complex data fusion, which is needed for feature space augmentation.